\documentclass{article} 

\usepackage{arabtex}
\usepackage{utf8}
\usepackage{iclr2020_conference,times}
\usepackage{url}

\usepackage{hyperref}

\usepackage{graphicx}

\title{Using Punkt for Sentence Segmentation in non-Latin Scripts: Experiments on Kurdish (Sorani) Texts}

\author{Roshna Omer Abdulrahman, Hossein Hassani \\
Department of Computer Science and Engineering\\
University of Kurdistan Hewl\^er
\\Kurdistan Region - Iraq\\
\texttt{\{roshna.abdulrahman,hosseinh\}@ukh.edu.krd} \\
}

\iclrfinalcopy 
\setcode{utf8}
\begin{document}

\maketitle

\begin{abstract}
Segmentation is a fundamental step for most Natural Language Processing tasks. The Kurdish language is a multi-dialect, under-resourced language which is written in different scripts. The lack of various segmented corpora is one of the major bottlenecks in Kurdish language processing. We used \textit{Punkt}, an unsupervised machine learning method, to segment a Kurdish corpus of Sorani dialect, written in Persian-Arabic script. According to the literature, studies on using \textit{Punkt} on non-Latin data are scanty. In our experiment, we achieved an F1 score of 91.10\% and had an Error Rate of 16.32\%. The high Error Rate is mainly due to the situation of abbreviations in Kurdish and partly because of ordinal numerals. The data is publicly available at \url{https://github.com/KurdishBLARK/KTC-Segmented} for non-commercial use under the CC BY-NC-SA 4.0 licence.
\end{abstract}


\section{Introduction}
\label{Intro}

Kurdish is a multi-dialect, Indo-European language which is written in several scripts and spoken by approximately 30 million people in different countries  \citep{hassani2018blark}. It is considered an under-resourced language, particularly due to the lack of digital sources for language processing \citep{ahmadi2019lex}.

\par We use \textit{Punkt} \citep{kiss2006unsupervised}, an unsupervised machine learning approach, to segment Kurdish Textbook Corpus - KTC \citep{abdulrahman2019developing}, which is in Sorani dialect, written in Persian-Arabic script. The Sentence Boundary Detection (SBD) has some issues in Sorani \citep{esmaili2012challenges}, and to the best of our knowledge, currently, there is no publicly available segmented corpus for Sorani.

We show that Punkt performs appropriately on non-Latin scripts, which suggests that it could be used to expedite the segmentation process not only for Kurdish texts in Persian-Arabic but also for languages that hold a similar situation. 

\par The rest of this paper is organized as follows. Section 2 presents the segmentation method. Section 3 presents the experiments, results, and discussion. Finally, Section 4 concludes the paper and suggests future work on the topic.

\section{ Method}
Three main approaches to SBD exist, rule-based, supervised, and unsupervised machine learning \citep{read2012sentence}. Through using \textit{Punkt} on KTC, we take the latter approach because it does not require an annotated corpus.

KTC includes 12 educational subjects of K-12 textbooks containing 693,800 tokens (110,297 types). We divide the corpus into a 90-10\% for the development (dev) and test set, respectively. The division covers each subject per level. We train \textit{Punkt} sentence segmenter on the corpus's dev set. 

To make the segmentation task more accurate, in addition to the traditional definition, we assume a sentence also could be a series of words (tokens) on a single line that is not followed by an end of sentence punctuation, such as \textbf{.}, \textbf{?}, and \textbf{!}. For instance, book title pages do not end in punctuation. We also identify abbreviations to let \textit{Punkt} do the SBD with less ambiguity. 
\par We test our approach on the test set manually by separating 10\% of the newly segmented sentences. We evaluate the performance measures, precision, recall, error rate, and F1 score.

\section{Results and Discussion}

\begin{figure}[h]
\begin{center}
\includegraphics[width=10cm, height=4cm,keepaspectratio]{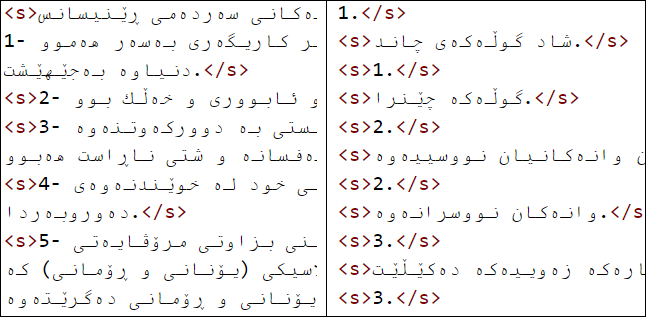}
\end{center}
\caption{Ordinal numerals sentence segmentation example.}
	\label{SegExamplef}
\end{figure}

We tokenized the dev set using NLTK's \citep{bird2009natural} word tokenizer, and trained a custom \textit{Punkt} sentence segmenter on the dev set. We fed the tokenized test set into the custom sentence segmenter and saved each file as an XML file of sentences tagged with the start and the end of sentence tags <s> and </s>, respectively (see Figure \ref{SegExamplef}).

\par We tested the sentence segmenter on the test set, and manually checked it by adding the attribute ``type'' to each segmented sentence with the values: \textit{tp} (true positive), \textit{tn} (true negative), \textit{fp} (false positive), and \textit{fn} (false negative).%

\par We observed that abbreviations\footnote{Abbreviations in Kurdish, particularly in texts written in Persian-Arabic script, are far less common compared to acronyms \citep{centralkurdishstyleguide2020}.} such as {({\<د. >}} Dr.), as well as ordinal numbers, cause a sentence to be split. As Figure~\ref{SegExamplef} shows, the punctuation mark after the numbers vary, it changes from a dot to a dash, and an underscore (\textit{.},\textit{-},\textit{\_}), which confuses the segmenter and creates more false positives.

\par We tested the same test set with five  abbreviation accounted for ({{\<د. >}} Dr.; {{\<پ. >}} professor; {{\<د.خ >}} S.W; {{\<م. >}} teacher; {\<پ‌.‌ز >} A.D). We obtained new results that showed identifying abbreviations would enhance the outcome. When only using \textit{Punk} without parameters, Precision (80.64\%) was lower compared to (83.67\%), while recall stayed unchanged at (100\%). The error rate difference was discernible; it dropped from (19.35\%) to (16.32\%). Finally, the F1 score was increased from (89.28\%) to (91.10\%)

\section{Conclusion}
We segmented Kurdish Sorani text into sentences using \textit{Punkt}, which follows an unsupervised machine learning approach. The experiments showed an F1 score of 91.10\% and an Error Rate of 16.32\% when giving abbreviations as parameters, but we received 19.35\% accuracy otherwise. The data set could be considered as part of the Kurdish BLARK. 

In the future, we are interested in applying other methods and, where is applicable, in merging the current one with rule-based methods to improve the accuracy of the approach.


\bibliography{iclr2020_conference}
\bibliographystyle{iclr2020_conference}


\end{document}